\documentclass[letterpaper]{article} 
\usepackage{aaai25}  
\usepackage{times}  
\usepackage{helvet}  
\usepackage{courier}  
\usepackage[hyphens]{url}  
\usepackage{graphicx} 
\usepackage{natbib}  
\usepackage{caption} 
\usepackage{algorithm}
\usepackage{algorithmicx}
\usepackage{algpseudocode} 
\usepackage{amsmath}
\usepackage{amsfonts}
\usepackage{newfloat}
\usepackage{listings}
\DeclareCaptionStyle{ruled}{labelfont=normalfont,labelsep=colon,strut=off} 
\floatstyle{ruled}
\newfloat{listing}{tb}{lst}{}
\floatname{listing}{Listing}
\makeatletter
\renewcommand{\fnum@algorithm}{\textbf{Algorithm~\thealgorithm}}
\makeatother

\title{Targeted Adversarial Denoising Autoencoders (TADA) \\for Neural Time Series Filtration}
\author{
    Benjamin J. Choi\textsuperscript{\rm 1}, 
    Griffin Milsap\textsuperscript{\rm 2}, 
    Clara A. Scholl\textsuperscript{\rm 2}, 
    Francesco Tenore\textsuperscript{\rm 2},
    Mattson Ogg\textsuperscript{\rm 2} \\
}

\affiliations{
    \textsuperscript{\rm 1}Harvard John A. Paulson School of Engineering and Applied Sciences, Cambridge, MA, United States of America\\
    \textsuperscript{\rm 2}Johns Hopkins University Applied Physics Laboratory, Laurel, MD, United States of America\\
    \footnotesize benchoi@college.harvard.edu, griffin.milsap@jhuapl.edu, clara.scholl@jhuapl.edu, francesco.tenore@jhuapl.edu,
    mattson.ogg@jhuapl.edu
}

\begin{document}

\maketitle

\begin{abstract}
Current machine learning (ML)-based algorithms for filtering electroencephalography (EEG) time series data face challenges related to cumbersome training times, regularization, and accurate reconstruction. To address these shortcomings, we present an ML filtration algorithm driven by a logistic covariance-targeted adversarial denoising autoencoder (TADA). We hypothesize that the expressivity of a targeted, correlation-driven convolutional autoencoder will enable effective time series filtration while minimizing compute requirements (e.g., runtime, model size). Furthermore, we expect that adversarial training with covariance rescaling will minimize signal degradation. To test this hypothesis, a TADA system prototype was trained and evaluated on the task of removing electromyographic (EMG) noise from EEG data in the EEGdenoiseNet dataset, which includes EMG and EEG data from 67 subjects. The TADA filter surpasses conventional signal filtration algorithms across quantitative metrics (Correlation Coefficient, Temporal RRMSE, Spectral RRMSE), and performs competitively against other deep learning architectures at a reduced model size of less than 400,000 trainable parameters. Further experimentation will be necessary to assess the viability of TADA on a wider range of deployment cases. 
\end{abstract}

%

\section{Introduction}

\begin{figure*}[t]
    \centering
    \includegraphics[width=0.83\textwidth]{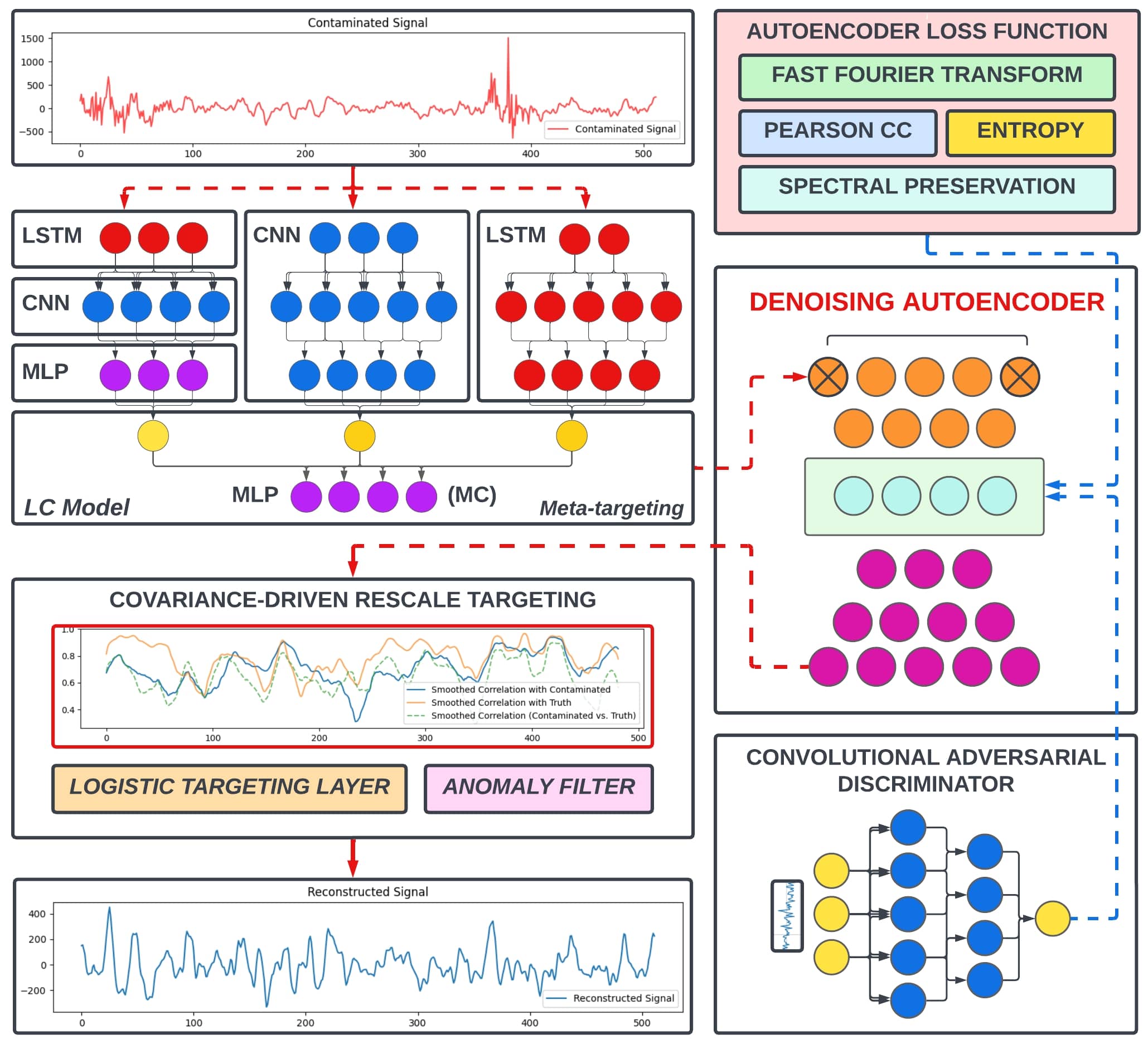}
    \caption{The architecture of the TADA system, with the targeting system depicted on the left and the adversarial autoencoder process on the right. The incoming signal is routed through the meta-targeting ensemble (upper left) before being passed through the autoencoder (right) and then through the covariance-driven rescale targeting function (lower left). The loss function (upper right) and the adversarial discriminator (lower right) assist in training. More extensive details are included in the Appendix.}
    \label{fig:figure1}
\end{figure*}

Developing effective time series filtration algorithms could help improve existing neural interfaces that rely on noisy time series data (Yadav et al., 2020; Linderman et al., 2008; Makeig et al., 2012). However, conventional neural time series filtration methods (canonical correlation analysis, or CCA; independent component analysis, or ICA, etc.) often fail to handle more complex single-channel time series denoising cases (Vergult et al., 2007; Chou et al., 2016; Vigario \& Oja, 2008). This is typically true, for example, with EMG interference in EEG data, which is a primary source of noise for neural interfacing applications (Stergiadis et al., 2022; Zhang et al., 2021a). Machine learning (ML)-based approaches, have made recent strides in handling single-channel time series denoising scenarios (Luo \& Mesgarani, 2019; Yin et al., 2023; Cui et al., 2024); ML-based time series filtration methods based on multilayer perceptrons (MLPs), convolutional neural networks (CNNs), recurrent neural networks (RNNs), and transformers have successively improved the state of the art (Zhang et al., 2021a; Cui et al., 2024; Yin et al., 2023). However, these published denoising approaches are compute-heavy, often involving large numbers of trainable parameters (and thus a large memory footprint) and long training times. These large compute footprints make deployment in real-time neural interfacing applications cumbersome (Justus et al., 2018; Cui et al., 2024; Yin et al., 2023; Buongiorno et al., 2021; Monga et al., 2021). Moreover, the inherent difficulties of EMG filtration continue to present challenges for deep learning (DL) models; complex patterns are harder to identify, so more exhaustive filtration is required to ensure a clean signal. In practice, this can lead to undesirable degradation (Cui et al., 2024; Yin et al., 2023; Justus et al., 2018). In short, existing ML-based filtration algorithms show promise, but can be weighed down by lengthy training times, large network sizes, and resulting issues with regularization, generalization, and application within dynamic real-world deployment cases (Monga et al., 2021; Purwins et al., 2019; Shrestha \& Mahmood, 2019). 

Autoencoders (AEs), which are designed to learn a latent representation of an incoming data stream, naturally lend themselves to potential applications in the world of blind source separation (BSS) and signal filtration. As AEs generally distill lower-complexity mappings of original data, traditional AEs are more lightweight in terms of parameter count and training time than other deep-learning architectures. This results in the potential for reduced overfitting and increased real-world applicability in dynamic environments requiring rapid iteration. However, while denoising AEs have achieved success in capturing and processing complex biological interference patterns (Ben Said et al., 2017; Dasan \& Gnanaraj, 2022; Rehman et al., 2018), performance remains modest compared to more expressive state-of-the-art (SOTA) DL approaches for neural time series denoising (Xiong et al., 2024; Rani et al., 2023; Dong et al., 2023). AEs typically incur temporal and spectral information degradation of neural signals (Shrestha \& Mahmood, 2019; Dasan \& Gnanaraj, 2022; Rehman et al., 2018). 

Generative adversarial (Goodfellow et al., 2020) training methods could potentially rectify such degradative behavior. By training a denoising AE in an adversarial regime, the output of an AE could be more nuanced—and thus, could better reproduce ground-truth EEG. Some initial studies (Rani et al., 2023; Dong et al., 2023) of adversarial denoising AEs have been reported and suggest some ability to improve ground-truth reconstruction. However, their performance (Dong et al., 2023) still lags SOTA DL in EEG-EMG denoising (Zhang et al., 2021a; Cui et al., 2024; Yin et al., 2023). Nevertheless, adversarial AEs appear to outperform their non-adversarial counterparts by essentially offloading some expressivity in the training process to an ultimately-discarded discriminator—rendering adversarial training a potentially useful tool when building denoising AEs for neural signals (Dong et al., 2023).

Our proposed model is situated in the aforementioned lineage of work on adversarial autoencoders. More specifically, to rectify the shortcomings of current AE denoising methods while leveraging their beneficial (lightweight, regularized) qualities, we propose a two-tiered targeting system to elevate traditional AE approaches to EEG denoising. We use the term “targeting” to refer to methods beyond the central AE model that externally optimize system inputs and outputs. Specifically, we employ two targeting solutions: (1) an initial ML-driven meta-targeting layer to selectively determine the requisite level of filtration, allowing our model to function without in-depth a priori knowledge of the incoming BSS problem contamination level, and (2) a post-AE logistic covariance-driven targeting method to calibrate an optimal RRMSE-preserving signal map. The first meta-targeting layer expands upon the conventional concept of a pre-filter interference check by selectively calibrating the model. To rectify suboptimal detection performance given the importance of signal-to-noise ratio (SNR) range flexibility, we apply a combined LSTM-CNN (LC) ensemble architecture (Choi \& Liu, 2025) to ensure broader coverage of potential interference cases. In the second targeting approach (logistic covariance-targeting), we minimize potential degradation by introducing a new weighted signal reconstruction method. Adversarial training is used to ensure the autoencoder output adheres to the characteristics of authentic EEG data. Further details are included in the ensuing sections and the Appendix.

\section{Methods}

The TADA system comprises three components: (1) the targeted denoising autoencoder, (2) adversarial training, and (3) a final logistic covariance targeting layer. Together, these address the need for a selective, expressive, non-destructive method for denoising neural time series data.

We surveyed numerous publicly available EEG time series datasets (Koelstra et al., 2011; Van Veen et al., 2019; Kaya et al., 2018; Zhang et al., 2021b) to study the proposed architecture, taking into account the scope of available data, number and length of individual segments (or channels), availability of a potential interference source, and the ability for ground truth verification. After careful consideration, we selected the EEGdenoiseNet benchmark dataset to train and evaluate the TADA system. EEGdenoiseNet was chosen due to its comprehensive scale (incorporating data from five separate studies), optimization for ML artifact removal training, and inclusion of ground-truth data—qualities favorable to those of other open-source options for the purpose of testing our hypotheses. The dataset provides “4514 clean EEG segments” and “5598 muscular artifact segments” at a 256 Hz digital sampling rate; the two-second segments enable “users to synthesize contaminated EEG segments with the ground-truth clean EEG” (Zhang et al., 2021b). Furthermore, since EEGdenoiseNet is a well-studied dataset, there are numerous denoising algorithms we could evaluate our approach against (Zhang et al., 2021a; Cui et al., 2024; Yin et al., 2023; Zhang et al., 2021b). From the EEGdenoiseNet dataset we synthesized three EEG-EMG training mixtures and three test mixtures at SNRs of -7 dB, -2.5, and 2 dB. A T4 GPU with an estimated 65 TFLOPS of FP16 precision was used for training and is referenced throughout.

We refer the readers to the original EEGdenoiseNet paper for complete details (Zhang et al., 2021b), but briefly, this corpus comprises mixtures of EEG signals with various known sources of noise (EMG artifacts) to support work on blind source separation problem for neural data. Contaminated EEG signals were generated via semi-synthetic EEG and EMG mixtures with hyperparameter selection calibrated to correspond to test cases spanning a conventional -7 dB to 2 dB SNR range for EMG-EEG denoising (Zhang et al., 2021b). The employed semi-synthetic contamination method is commonplace in the neural BSS field (Zhang et al., 2021a; Cui et al., 2024; Yin et al., 2023; Zhang et al., 2021b) and well-validated as a proxy for real-world testing (Zhang et al., 2021a; Cui et al., 2024; Yin et al., 2023; Zhang et al., 2021b). The following subsections detail the individual subcomponents of the TADA system.

\subsection{The Targeted Denoising Autoencoder}

To develop the proposed meta-targeting model, this study implemented a modified LSTM-CNN (LC) ensemble network (based on Choi \& Liu, 2025) designed for exhaustive EEG-specific coverage across a wide variety of signal forms (see Appendix for details). The leveraged architecture was trained to predict SNRs across a conventional (Zhang et al., 2021b) -7 to 2 dB case range from 1,200 synthetically-contaminated samples with a partitioned 100-sample test case to align with subsequent later-stage testing environments. After hyperparameter tuning, the LC model was able to correctly classify the SNR of every case in the test set; recorded mean test loss was 4.86E-3 across the -7 to 2 dB range. (An ablation analysis determined that removing individual ensemble components resulted in imperfect SNR classification scores among our test cases, thereby justifying increases in model complexity.) The LC model’s ability to develop a virtually solved representation of the meta-targeting problem was not particularly surprising given the demonstrated model diversity and expressivity, the relative prominence of EMG artifacts, and past demonstrated detection performance (e.g., 98\% reported in Soroush et al., 2022). However, the LC model’s additional SNR targeting component combined with a virtually complete learned representation does offer new promise for tailored AE-implementation by offloading the expressivity inherent in SNR calibration to an upstream meta-targeter. Total LC training time over 100 epochs on a T4 GPU was 81.8 seconds.

Following LC completion, an initial denoising autoencoder (AE) was trained to filter EMG artifacts after LC meta-targeting. We implemented our AE model using a convolutional architecture (see Appendix for details) to encode important spatiotemporal patterns in the EEG data. This approach achieved auto-encoding performance (i.e., reconstruction errors) in line with previous work (Dong et al., 2023), around 0.7 in terms of CC, which we further improved via additional processing stages (see below). Note that we found this process to be sensitive to various architecture and hyper-parameter changes, often returning degradative outputs prior to parameter fitting and optimization.

\begin{figure*}[t]
    \centering
    \includegraphics[width=0.99\textwidth]{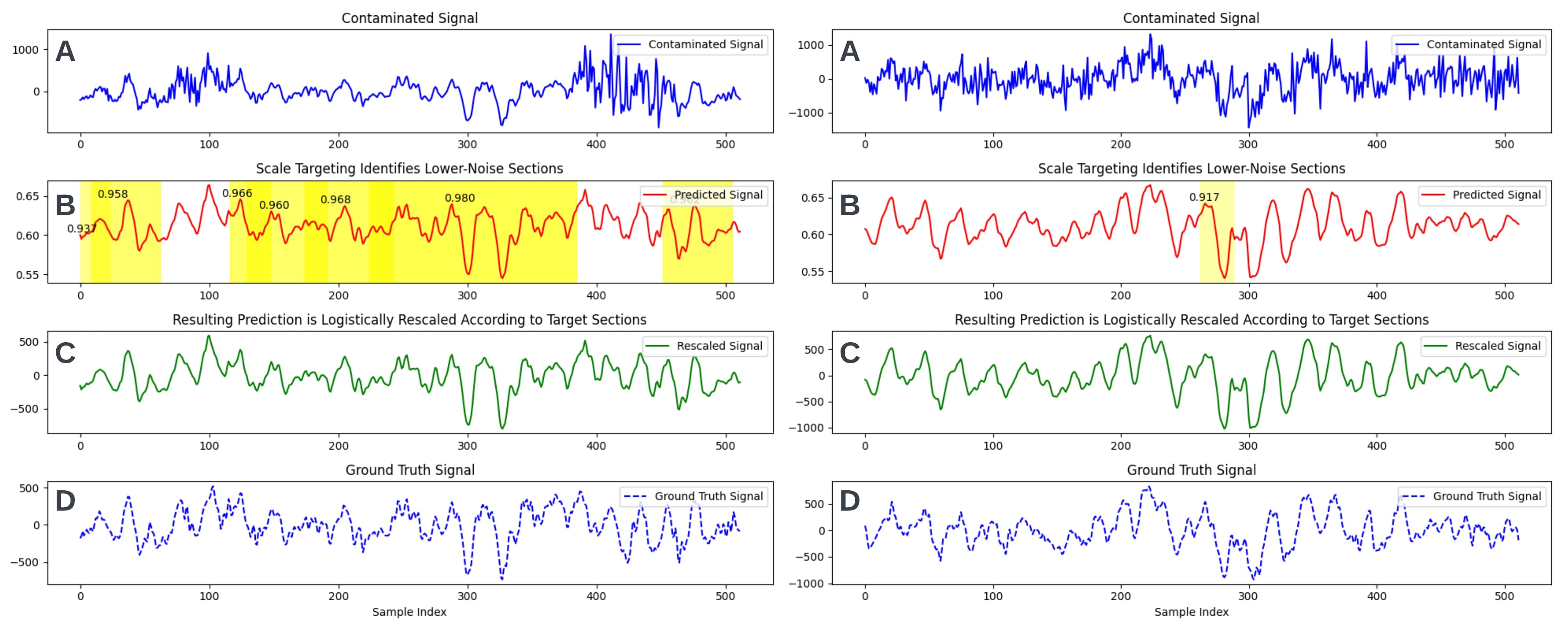}
    \caption{Two examples (one left, one right) of the logistic target rescaling workflow leveraging low noise segment identification and weighting. Contaminated signals (A) are routed through the autoencoder, and the subsequent scale targeting process (B) maps the AE output to the appropriate final range (C). Ground truth is depicted in the bottom panel (D) for reference. More precise formulations are included in the Appendix.}
    \label{fig:figure2}
\end{figure*}

\subsection{Loss Function and Adversarial Training}

To improve convolutional autoencoder performance, two additional steps were taken: (1) development of a custom loss function (see Appendix for details), and (2) an adversarial training approach. Rather than employing compute-heavy, MSE-driven optimization, we abstracted the objective of correlative reconstruction into a novel objective function. Temporal MSE preservation was outsourced to the subsequent logistic covariance scale targeting algorithm (see section 2.3). To address the previously delineated concerns with low-entropy reconstruction and spectral loss, we wrapped entropy-incentivization and spectral preservation metrics into the loss function in conjunction with the primary CC objective. By employing streamlined covariance-driven CC computation and Fast Fourier Transform power spectrum-based cross-comparison to ensure frequency domain similarity, we enabled both faster training time and expressivity throughout the convolutional autoencoder training process. This loss function and subsequent rescaling resulted in mean AE performance gains of 4.20\%, 13.9\%, and 17.1\% for CC, TRRMSE, and SRRMSE respectively. 

Performance was improved further via adversarial training. Leveraging the local self-similarity and translational invariance inherent in EEG, we trained a DL convolutional discriminator model to discern between AE-reconstructed and authentic ground truth EEG samples (see Appendix for details). The adversarial training process was designed to enforce more realism on the autoencoder output and filtration, with rescaling conducted before passing samples to the discriminator. Hyperparameter tuning was conducted during the adversarial training process to ensure competitive balance between the AE generator and DL convolutional discriminator. Respective losses were carefully monitored to ensure relative parity.

This modestly sized model was trained for five generator-discriminator cycles. In total, adversarial autoencoder training was completed in 57.7 seconds on a T4 GPU (resulting in an overall training time of 139.5 seconds when combined with the LC meta-targeting training run). These five cycles of adversarial autoencoder training led to mean improvements of 8.6\%, 14.5\%, and 14.6\% for CC, TRRMSE, and SRRMSE, respectively—testifying to the potential benefit of the discriminator at ensuring the denoised output adheres to authentic EEG characteristics. While the convolutional discriminator itself does not play a role in final system inferencing, model layer sizes were regulated to (1) ensure competitive parity with the autoencoder during adversarial training, and (2) enable faster training.

Crucially, when considering overall training runtime in a dynamic BSS environment, it is essential to consider both the combined length of the EEG and EMG segments used in (re-)training as well as the model training run itself. For a BSS model to successfully retrain itself in a live streaming environment, the total effective training time is not merely the length of the training run, but rather the maximum of both the training run and the total duration of training data needed to retrain the model. For the TADA system, training data was restricted to a randomly selected group of 300 EEG segments contaminated with corresponding high-variance EMG artifacts across the SNR spectrum—totaling 600s in combined duration. Thus, the effective training time of the TADA system in a real-time environment is roughly 10 minutes; the TADA model can be initially trained (e.g, on the EEGdenoiseNet data) in one-fourth of this time, but at least 10 minutes worth of in-domain data are required to support a successful training run.

\subsection{Covariance-Driven Logistic Scale Targeting}

\begin{figure*}[t]
    \centering
    \includegraphics[width=0.9\textwidth]{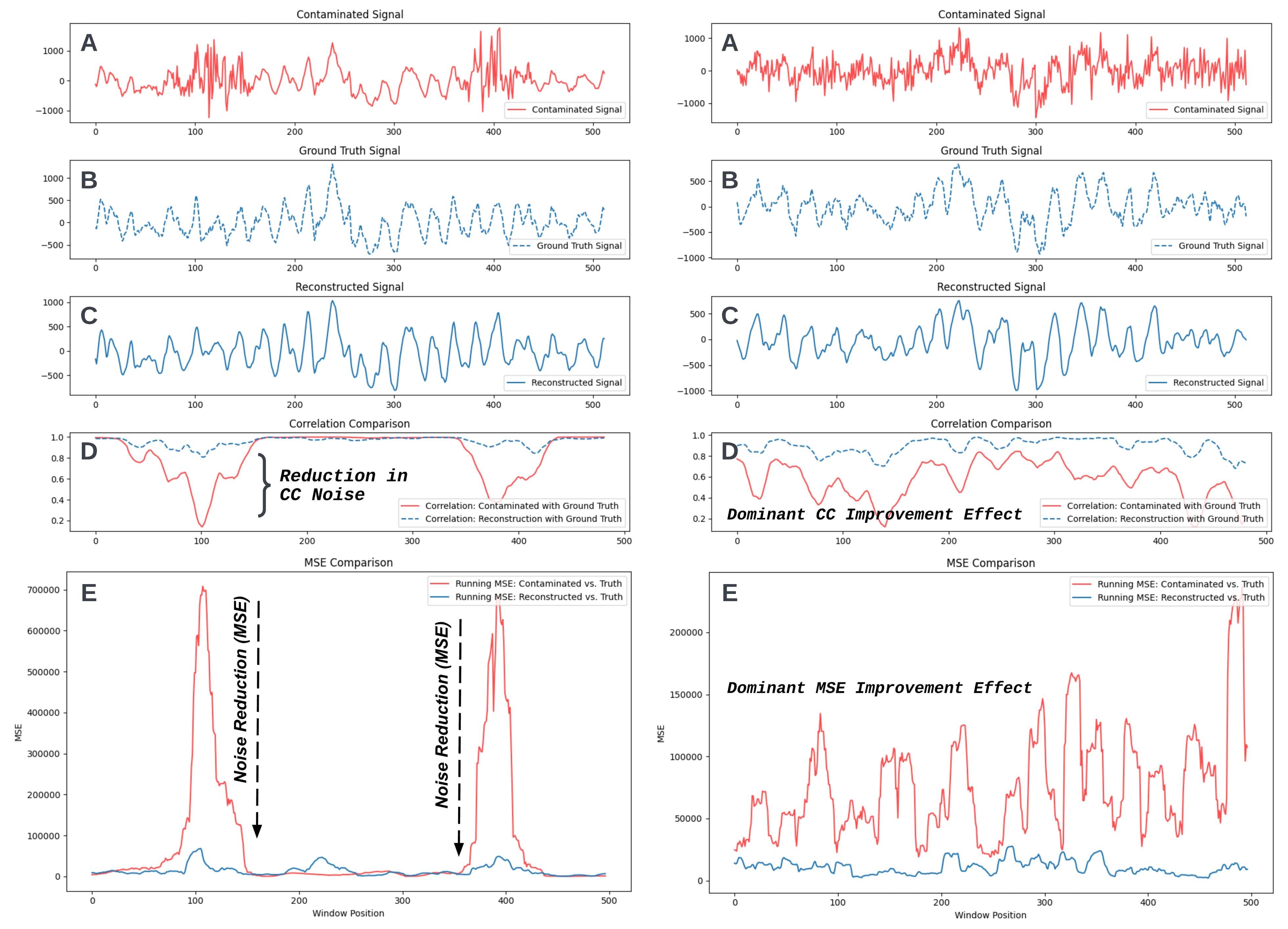}
    \caption{Two examples of TADA filtration demonstrating system noise reduction capabilities (spike artifact on the left; continuous artifact on the right). Panels (A), (B), and (C) depict contaminated, ground truth, and the TADA-reconstructed signal, respectively. TADA time series filtration yields improved correlation with ground truth (D) and reduced mean squared error (E).}
    \label{fig:figure3}
\end{figure*}

Following completion of GAN training, further algorithmic efforts to reduce temporal and spectral RRMSE loss were performed via the development of a novel covariance-driven logistic scale targeting algorithm (see Appendix for details). 

The scale targeting algorithm is the nucleus of the TADA signal reconstruction process and was built to enable elegant generalization: the algorithm’s fundamental mechanics are theory-driven, not use-case-specific. Optimized for streamlined, memory-efficient, low-latency performance, the algorithm is designed to map the high-correlation AE output—which is passed out of the AE in unscaled form—into an optimal MSE-reducing reconstruction. Rather than take a blind compute-heavy approach, the algorithm attempts to determine sections of the original signal with low noise by analyzing the FIR-passed running correlation between the original and AE-suggested signal. Since the AE-suggested signal is intended to mimic pristine ground-truth EEG, sections of the original contaminated signal that post strong covariance with the AE suggestion can be heuristically inferred to have lower noise. 

The resulting low-noise target sites are then logistically weighted to determine a composite overall target site representation; the scale targeting algorithm subsequently maps the AE output to the final optimal MSE reconstruction based on the composite offset and scale. Thus, while concurrent CC and MSE optimization requires significant compute and system expressivity, by streamlining the process via the AE-to-scale-target chain, our sequential fine-tuned ML system enables effective low-parameter signal filtration (as evident in the aforementioned ablation data).

Following scale targeting algorithmic development, several augmenting mechanisms were developed to handle potential edge cases. The first edge case is intuitive: if there are no high-covariance segments in the original signal satisfying the lowest system threshold bound. Experimentally, iteratively lowering the correlation threshold led to worse MSE performance—the system ended up rescaling based on high noise sites, leading to AE-output maps with sharp divergence from the underlying ground truth. Instead of overextending the targeting algorithm in the hopes of addressing the most underdetermined instance of a fundamentally underdetermined problem, we achieved superior results on a 300-sample test set by reverting to a more conventional strategy of mapping edge cases to a dataset-average offset and amplitude ratio. In the final test set, this handling mechanism was invoked at a frequency of 0\% at high SNR (2 dB), 8.0\% at mid SNR (-2.5 dB), and 14.0\% at low SNR (-7 dB). 
	
The second edge case that emerged during validation is specific to the developed scale targeting algorithm: that of isolated stochastic noise alignment. Specifically, a phenomenon emerged during validation where oscillating EMG artifacts in the original signal could—by random chance—happen to briefly align with the autoencoder output. While this effect is mitigated in signals containing other low-noise target sites off which to base the final transformation, if occurring coincidentally in isolation, an unlucky spurious correlation could drastically shift the scaled output and yield suboptimal MSE. Fortunately, this anomalous event has an identifiable signature—the resulting output from isolated stochastic noise alignment is typically mapped to a high-magnitude, extreme-amplitude range—and thus samples satisfying this criterion can be generically rescaled in accordance with the previous edge case handling protocol. In the final test set, this second edge case conditional was invoked at a frequency of 0\% at high SNR (2 dB), 1.0\% at mid SNR (-2.5 dB), and 8.0\% at low SNR (-7 dB). While the performance effect was minimal at mid and high SNR, RRMSE improvements on the order of a 25\% reduction in error were observed following edge case handling implementation at the low end of the SNR range.

\section{Results}

\begin{table*}[t]
\centering
\renewcommand{\arraystretch}{1.2} 
\fontsize{10}{12}\selectfont      
\setlength{\tabcolsep}{8mm}       
\begin{tabular}{lccc}
\hline\hline
\textbf{Metric} & \textbf{Low SNR (-7 dB)} & \textbf{Mid SNR (-2.5 dB)} & \textbf{High SNR (2 dB)} \\
\hline
Correlation Coefficient & 0.6856 & 0.8468 & 0.9447 \\
\hline
Temporal RRMSE & 0.8156 & 0.5664 & 0.3357 \\
\hline
Spectral RRMSE & 0.8997 & 0.5706 & 0.2683 \\
\hline\hline
\end{tabular}
\caption{TADA system performance summary. SNR refers to the signal-to-noise ratio of the contaminated input signal.}
\label{tab:tada_performance_summary}
\end{table*}

\begin{table*}
\centering
\renewcommand{\arraystretch}{1.2} 
\fontsize{10}{12}\selectfont      
\setlength{\tabcolsep}{4.1mm}       
\begin{tabular}{lccccl}
\hline\hline
\textbf{Model} & \textbf{C-T-S (-7 dB)} & \textbf{C-T-S (-2.5 dB)} & \textbf{C-T-S (2 dB)} & \textbf{Est. Parameters} \\
\hline
Novel CNN \citep{Zhang2021a} & 0.69-0.72-0.65 & 0.89-0.40-0.45 & 0.92-0.33-0.30 & 58.7M \\
\hline
EEGIFNet \citep{Cui2024} & unknown & 0.91-0.40-unk & 0.95-0.32-unk & 5.9M \\
\hline
GCTNet \citep{Yin2023} & unknown & 0.93-0.31-unk & 0.94-0.28-unk & $\sim$10M \\
\hline
\textbf{TADA (Ours)} & 0.69-0.82-0.90 & 0.85-0.57-0.57 & 0.94-0.34-0.27 & 393K+ \\
\hline\hline
\end{tabular}
\caption{Comparison with state-of-the-art ML denoising algorithms. C-T-S denotes CC, TRRMSE, SRRMSE, respectively. All values best extrapolated from provided code (or documentation). Unreported values are listed as unknown (unk).}
\label{tab:comparison_sota_dl}
\end{table*}

We conducted quantitative assessments analyzing TADA performance in aggregate on the test partition. Evaluation on the test partition was conducted across the aforementioned -7, -2.5, and 2 dB SNR cases. In accordance with field convention (Zhang et al., 2021b), analyses of Pearson correlation coefficient (CC), temporal relative root mean squared error (TRRMSE), and spectral relative root mean squared error (SRRMSE) metrics were computed to allow for some comparisons with existing literature. These three metrics describe correlation with ground truth, error in the temporal domain, and error in the frequency domain, respectively. The overall performance of our TADA system is presented in Table 1; Table 2 summarizes TADA’s performance relative to published major EEGdenoiseNet benchmarks (Zhang et al., 2021b). TADA is competitive with state-of-the-art performance on EEGdenoiseNet on the tested cases despite a much smaller model footprint.

Moreover, model latency, size in terms of trainable parameters (linked with space efficiency), and overall training time were assessed in accordance with the fundamental objectives of developing a streamlined, flexible system viable for real-world application in dynamic deployment cases. Overall mean TADA system latency was assessed at 1.47 milliseconds on a T4 GPU; the mean latency breakdown between initial LC meta-targeting, autoencoder inferencing, and logistic covariance-driven rescaling is 71.7\%, 25.9\%, and 2.4\% respectively.

Beyond computing major aggregate CC, TRRMSE, SRRMSE metrics (e.g., mean, median, etc.), we also conducted distribution analyses across the test set to assess TADA performance in wholesale fashion akin to past precedent in literature (Yin et al., 2023; Zhang et al., 2021b). These distribution analyses allowed for further quantitative exploration of edge case handling, future comparative algorithmic assessments, potential performance discontinuities, and visualization of worst-case performance scenarios across a wide SNR range. The Appendix provides a graphical summary of performance distribution across the partitioned test set.

We also conducted qualitative evaluations of the demonstrated TADA system across the tested deployment cases. Instances of myoelectric contamination can be informally categorized into “continuous interference” (i.e., continuous lower-to-mid-amplitude-range artifacts across the entire segment) and “spike artifacts” (i.e., isolated high-amplitude artifacts originating from short, intense contraction). Qualitative evaluation of TADA filtration performance on these two distinct scenarios yielded promising results and notable improvements in terms of MSE and CC were observed in both situations. Figure 3 provides a visualization of system performance on test samples that depict these two different qualitative scenarios. 

Qualitative analyses were also conducted to understand TADA performance given certain extreme examples to assess the boundaries, potential, and limitations of the system. In terms of pure MSE and CC noise reduction, the largest performance improvements were observed at lower SNRs (e.g., -7 dB); individual reconstructive CCs exceeding 0.85 were observed at -7 dB initial SNR, speaking to the effectiveness of the adversarial convolutional autoencoder. Significant spectral and temporal MSE improvements were also observed at low and mid initial starting SNR, indicating both the streamlined efficacy of logistic covariance-driven rescaling and the general success of edge case handling mechanisms in conjunction with the autoencoder. In terms of MSE reduction, high starting SNRs led to samples with the lowest recorded TADA-driven improvement—this finding is not surprising given that a high starting SNR naturally caps future improvement via filtration. Notably, TADA at high starting SNRs did not exhibit degradative behavior; signal adjustments, in aggregate, were beneficial; the quantitative results (Tables 1 and 2) underscore TADA’s superior performance at high starting SNR levels. Examples of best- and worst-case filtration are illustrated in the Appendix.

Finally, we wanted to compare TADA’s performance to other SOTA EEG denoising algorithms (Zhang et al., 2021a; Cui et al., 2024; Yin et al., 2023). Note, however, that these studies sometimes employed different testing protocols (Cui et al., 2024; Yin et al., 2023), benchmarks, closed access data and code, and modifications of (or deviations from) the standard guidelines outlined in Zhang et al.’s original EEGdenoiseNet dataset report (Zhang et al., 2021b). This complicates perfect one-to-one comparison across all SOTA denoising algorithms, but reported results are displayed in Table 2 to the extent possible. Moreover, in light of the core DL signal processing challenges outlined in the Introduction (Zhang et al., 2021a; Justus et al., 2018; Sainath et al., 2012; Buongiorno et al., 2021; Monga et al., 2021; Purwins et al., 2019; Shrestha \& Mahmood, 2019), a model size comparison (Table 2, Est. Parameters) is also provided—TADA is expressly designed to be a streamlined, space-efficient filtration system enabling real-time processing, on-the-fly training and subsequent adaptation to dynamic real-world deployment cases.

\section{Discussion}

Compared to SOTA DL algorithms on EEG time series filtration, TADA ties Zhang et al.’s Novel CNN (Zhang et al., 2021b) in terms of lower bound SNR CC and exceeds the performance in (Zhang et al., 2021b) on upper bound spectral RRMSE, but currently performs worse than the top published DL algorithm on the remaining metrics across the SNR spectrum. In further distributional analyses (see Appendix), we observe that the single-segment performance discrepancy between TADA and other top DL models falls within a 90\% error interval (judged by approximate standard deviation) across all metrics—with the sole exception of TRRMSE at -2.5 dB, for which GCTNet’s performance is roughly 2.3 standard deviations above the TADA mean (Yin et al., 2023). Thus, across the -7 dB to 2 dB range of SNR environment cases, TADA’s performance may occasionally exceed top DL performance means, but overall falls short of the best existing DL models—falling particularly short compared to GCTNet at mid-range SNR. (More extensive information is contained below in the Appendix.)

An advantage of the TADA system is its reduced parameter count, especially in light of the advantages of streamlined, space-efficient DL models (Thompson et al., 2021). TADA’s rapid training time, coupled with its aforementioned competitive performance, offers great utility and flexibility in comparison with top published algorithms. At less than 10\% of the estimated size of its nearest published DL equivalent, TADA’s compact format opens up doors to flexibility, rapid real-time training (and updating), and a broader range of real-world deployment opportunities. By shirking a conventional compute-heavy DL approach in favor of outsourcing more complex DL tasks to tailored upstream targeting mechanisms (i.e., SNR inferencing via LC meta-targeting, MSE reduction via logistic covariance scale targeting), TADA was able to train in just an average of 139.49 seconds (on a T4 GPU with an estimated 65 TFLOPS of FP16 precision), with an 0.58:0.41 training time breakdown between LC meta-targeter and autoencoder. As previously mentioned, it is important to note that training time is hardware- and dataset-dependent and that TADA’s effective training time in an online streaming environment exceeds 139.49 seconds as the model requires at least 10 minutes worth of EEG training data to achieve the performance described in this study. (Note that the other DL denoising models presented in Table 2 were trained on EEG data exceeding 120 minutes in combined duration.) Long training time requirements inhibit DL algorithms from adjusting to new information and dynamically changing data streams, which is a particular challenge for DL models tailored to signal processing. We acknowledge that training times are inherently less comparable due to differing computational resources and the stochastic nature of ML optimization. However, estimated training cost can be roughly inferred by considering the number of adjustable parameters and the resulting size of the space their combinations occupy; a $>$90\% reduction in trainable parameters confers implied benefit in the realm of training burden. TADA’s lightweight structure opens many doors for real-world use: deployment in compute-restricted environments, real-time handling of high-variance EMG artifacts, live retraining during streaming scenarios, and possible edge device usage.

Beyond enabling new capabilities in the non-invasive neural interfacing space by addressing perennial signal quality challenges, this preliminary study demonstrates the effectiveness of our TADA system prototype and raises opportunities for multiple possible extensions. TADA’s generalizable, theory-driven architecture could enable other fruitful applications within the wider signal-processing world, particularly given the ubiquity of the BSS paradigm. Depending on the intended use case, further TADA model size and training time reductions could be achieved by (1) easing model performance constraints, especially if the target interface SNR falls in a narrower band than -7 to 2 dB, (2) scaling to larger training datasets, or (3) relaxing the virtual solvability requirement of the LC meta-targeter (see Appendix for more details). In addition, our TADA prototype functioned largely independent of conventional feature engineering (e.g., initial bandpass filtration)—pairing TADA with traditional signal processing techniques could further refine performance. Though our initial demonstration of TADA (and the EEGdenoiseNet corpus) is largely single-channel in application, the system is also easily generalizable to multi-channel denoising situations—either by independently applying TADA to each channel, or by collectively leveraging correlative interchannel patterns by replacing the single-channel covariance computation in the rescale targeting algorithm with a deflation-accelerated linear multi-channel analog.
    
The results reported here are promising, but further studies will be necessary to validate TADA performance across a wider range of artifact types (e.g., beyond high-variance EMG artifacts in the -7 to 2 dB SNR range). Similarly, while outside of the scope of this study, more extensive studies involving human participants in diverse neural interfacing use cases will be important to establish real-world applicability.

\nocite{*}
\bibliography{aaai25}

\section{Appendix}

\subsection{Algorithmic Details}

\begin{algorithm}
\caption{Covariance-Driven Logistic Scale Targeting}
\label{alg:ScaleTargeting}
\begin{algorithmic}
\Function{ScaleTargeting}{$\mathbf{A}, \mathbf{B}, \tau, w$}
    \State $L \gets \min(\text{len}(\mathbf{A}), \text{len}(\mathbf{B}))$
    \State $\mathbf{A} \gets \mathbf{A}_{:L}$
    \State $\mathbf{B} \gets \mathbf{B}_{:L}$
    \State $\mathbf{R} \gets$ empty list 
    \For{$i = 0$ \textbf{to} $L - w$}
        \State $r \gets \text{corr}(\mathbf{A}_{i:i+w}, \mathbf{B}_{i:i+w})$
        \State append $r$ to $\mathbf{R}$
    \EndFor
    \State Initialize $\omega_{\text{c}}, \omega_{\text{p}}, \Omega$ to $0$
    \For{$i = 0$ \textbf{to} $\text{len}(\mathbf{R}) - 1$}
        \If{$\mathbf{R}[i] > \tau$}
            \State $\omega \gets \frac{1}{1 + \exp(-20(\mathbf{R}[i] - \tau))}$ 
            \State $\Sigma_{\text{c}} \gets \text{mean}(\mathbf{B}_{i:i+w})$
            \State $\Sigma_{\text{p}} \gets \text{mean}(\mathbf{A}_{i:i+w})$
            \State $\omega_{\text{c}} \gets \omega_{\text{c}} + \Sigma_{\text{c}} \cdot \omega$
            \State $\omega_{\text{p}} \gets \omega_{\text{p}} + \Sigma_{\text{p}} \cdot \omega$
            \State $\Omega \gets \Omega + \omega$
        \EndIf
    \EndFor
    \If{$\Omega = 0$}
        \State $\mathbf{A}_{\text{rescaled}} \gets \Call{StandardRescale}{\mathbf{A}}$
    \Else
        \State $\mu_{\text{c}} \gets \frac{\omega_{\text{c}}}{\Omega}$
        \State $\mu_{\text{p}} \gets \frac{\omega_{\text{p}}}{\Omega}$
        \State $\sigma_{\text{c}}^2 \gets \text{Var}\left(\{\mathbf{B}_{i:i+w} \mid \mathbf{R}[i] > \tau\}\right)$
        \State $\sigma_{\text{p}}^2 \gets \text{Var}\left(\{\mathbf{A}_{i:i+w} \mid \mathbf{R}[i] > \tau\}\right)$
        \State $\mathbf{A}_{\text{centered}} \gets \mathbf{A} - \mu_{\text{p}}$
        \State $\mathbf{A}_{\text{rescaled}} \gets \mathbf{A}_{\text{centered}} \cdot \left(\frac{\sigma_{\text{c}}}{\sigma_{\text{p}}}\right) + \mu_{\text{c}}$
        \State $\mathbf{A}_{\text{rescaled}} \gets \Call{AnomalyFiltration}{\mathbf{A}_{\text{rescaled}}, \mathbf{B}}$
    \EndIf
    \State \Return{$\mathbf{A}_{\text{rescaled}}$}
\EndFunction
\end{algorithmic}
\end{algorithm}

Algorithm~\ref{alg:ScaleTargeting} outlines the steps of the Covariance-Driven Logistic Scale Targeting algorithm, a core component component of the TADA system's denoising pipeline. This algorithm processes the autoencoder's output in conjunction with the original contaminated EEG signal to compute optimal scaling and offset parameters that minimize the Relative Root Mean Squared Error (RRMSE) while preserving the Pearson Correlation Coefficient (CC) between the denoised and ground truth signals. The algorithm operates by sliding a window across the original signal to compute running correlations, using high correlation between the filtered signal and the original signal as a heuristic for low noise. This targeted approach ensures that the denoised signal closely aligns with the ground truth in regions of high fidelity while effectively suppressing noisy segments.

To further enhance the reliability of the logistic scale targeting algorithm, the TADA system incorporates specialized mechanisms to address potential edge cases, as discussed:

\begin{enumerate}
    \item \textbf{Absence of high correlation segments}: In instances where no segments within the EEG signal exceed the predefined correlation threshold $\tau$, the algorithm defaults to applying a standard rescaling. This approach prevents the system from making arbitrary or detrimental adjustments in the absence of reliable low-noise regions.
    \item \textbf{Isolated stochastic noise alignment}: Occasionally, high-amplitude EMG artifacts may inadvertently align with the autoencoder's output within a sliding window, resulting in spurious high correlations. To mitigate the impact of such anomalies, the algorithm includes an anomaly filtration step that detects and corrects extreme deviations in the rescaled signal. This ensures that isolated noise alignments and spurious rescaling do not compromise the denoised EEG signal.
\end{enumerate}

\begin{figure}
    \centering
    \includegraphics[width=\columnwidth]{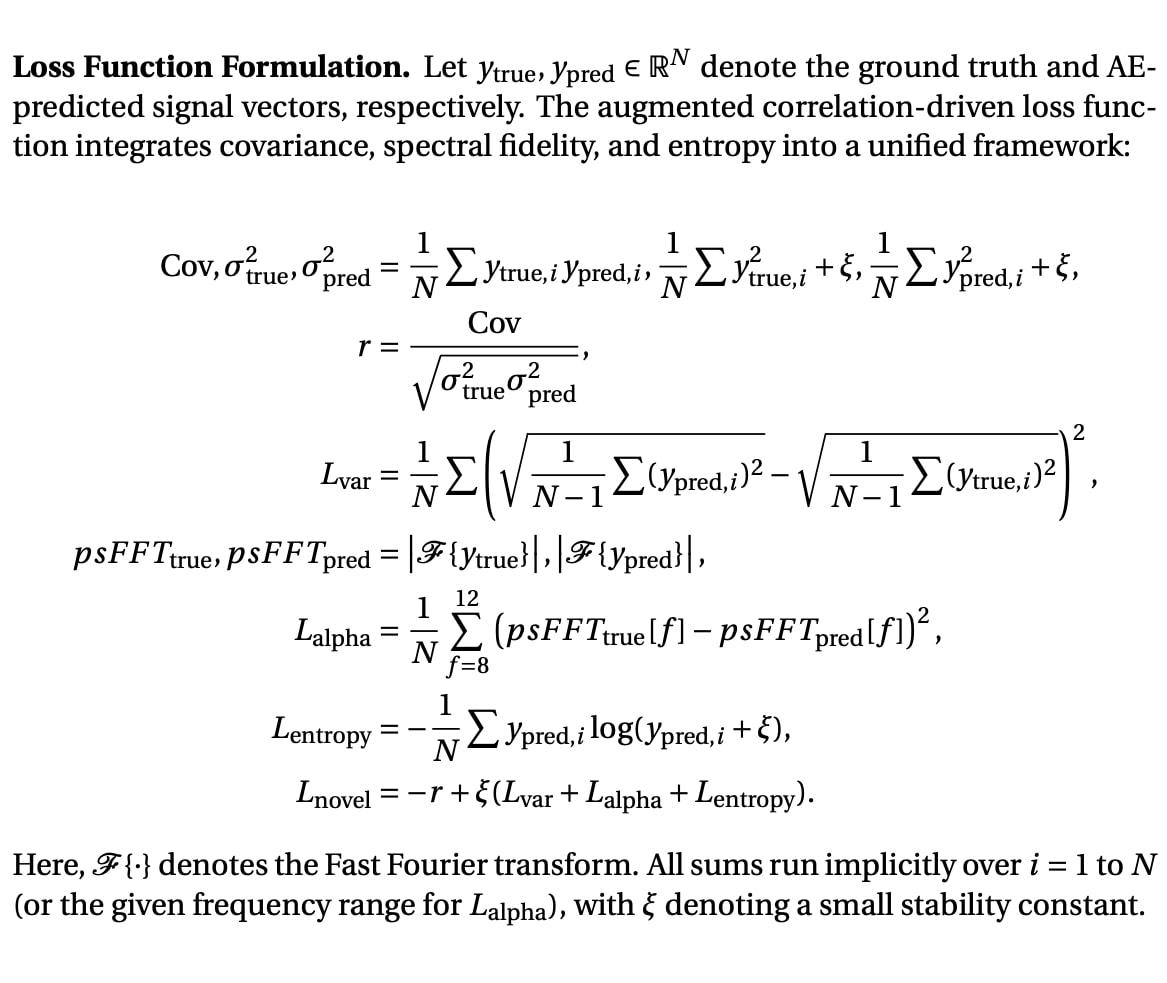}
    \caption{The correlation-driven autoencoder loss function.}
    \label{fig:loss}
\end{figure}

\begin{figure*}[t]
    \centering
    \includegraphics[width=0.99\textwidth]{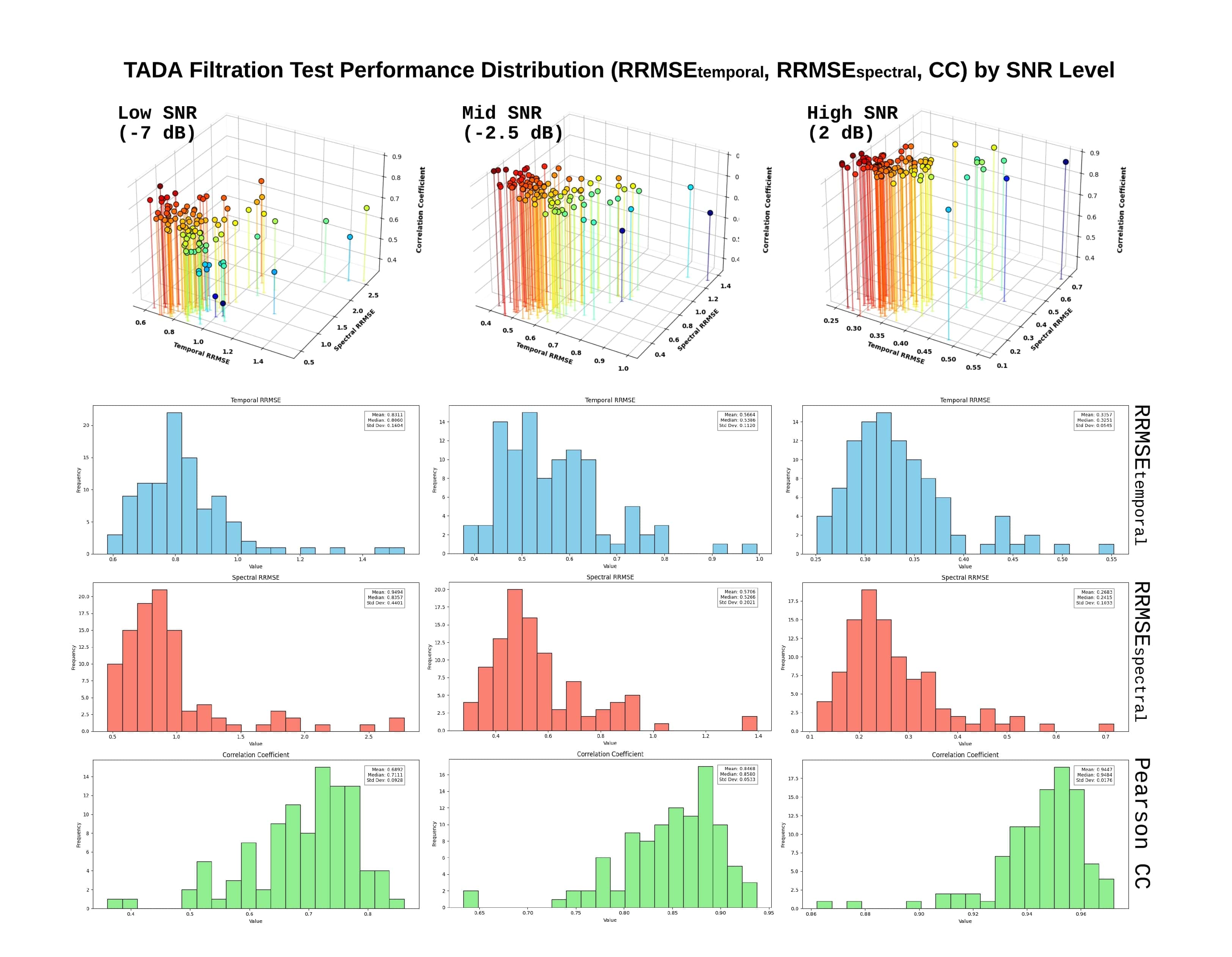}
    \caption{CC, TRRMSE, and SRRMSE test performance distribution on the -7 dB, -2.5 dB, and 2 dB test cases. The upper panels depict performance scatterplots of the individual test samples, with CC, TRRMSE, and SRRMSE denoted on the three axes. The lower panels depict histograms (from top to bottom) of TRRMSE, SRRMSE, and CC, respectively.}
    \label{fig:distribution}
\end{figure*}

In conjunction with the logistic scale targeting, we also incorporate a new CC-based loss function (Figure 4). As detailed in the methods section, instead of focusing on compute-heavy MSE-driven optimization, we abstracted the objective of correlative reconstruction into a streamlined AE loss function (while outsourcing temporal MSE preservation to the subsequent logistic scale targeting algorithm).

\begin{figure*}[htb]
    \centering
    \includegraphics[width=0.9\textwidth]{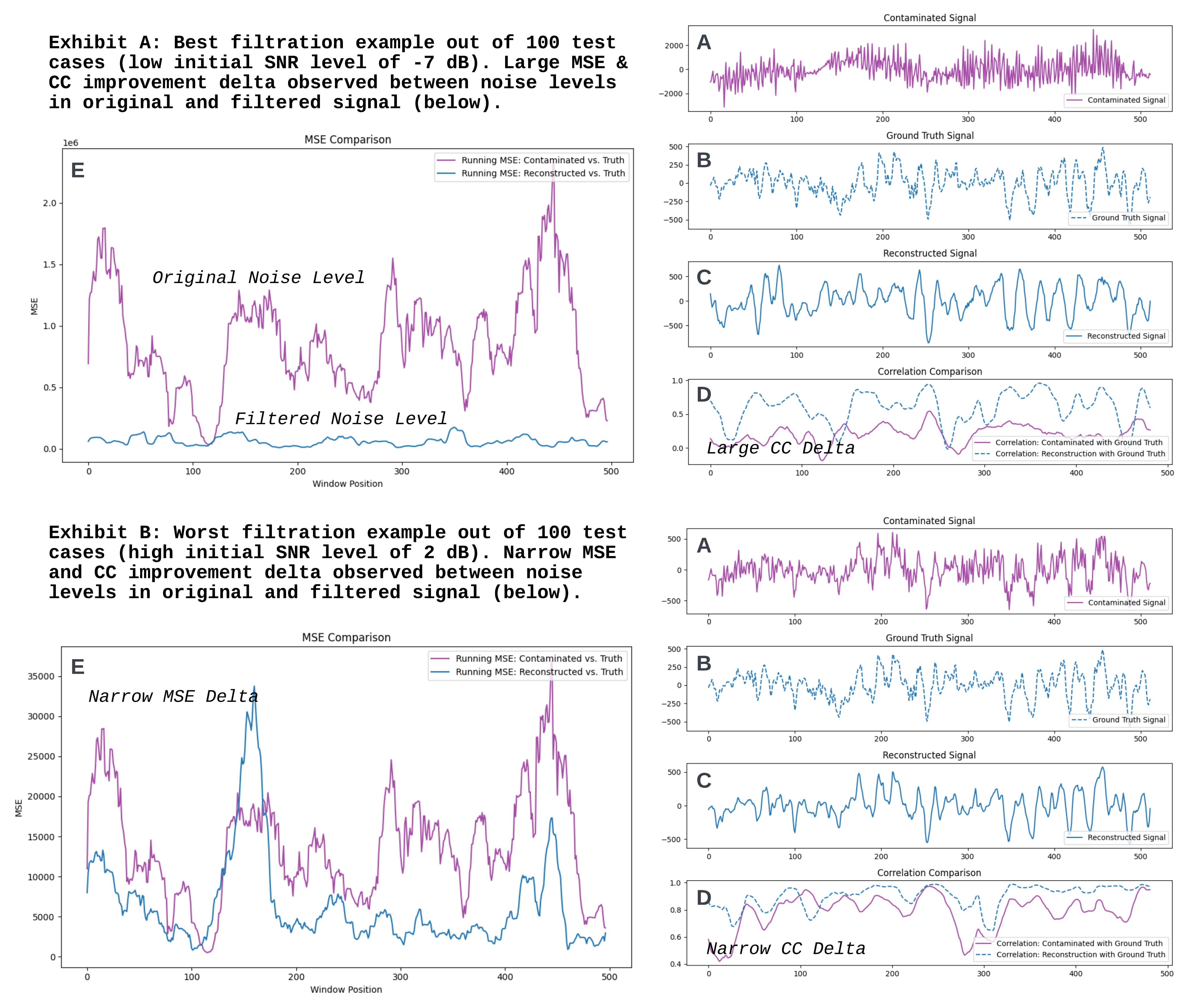}
    \caption{Best- and worst-case test set examples of TADA filtration. The right panels depict (A) contaminated, (B) ground truth, (C) TADA-reconstructed signal, and (D) pre- vs. post-TADA correlation with ground truth. The left panels (E) depict MSE error in the original (pre-TADA) vs. filtered (post-TADA) signals.}
    \label{fig:cases}
\end{figure*}

\subsection{Further Results}

\begin{figure*}[t]
    \centering
    \includegraphics[width=0.9\textwidth]{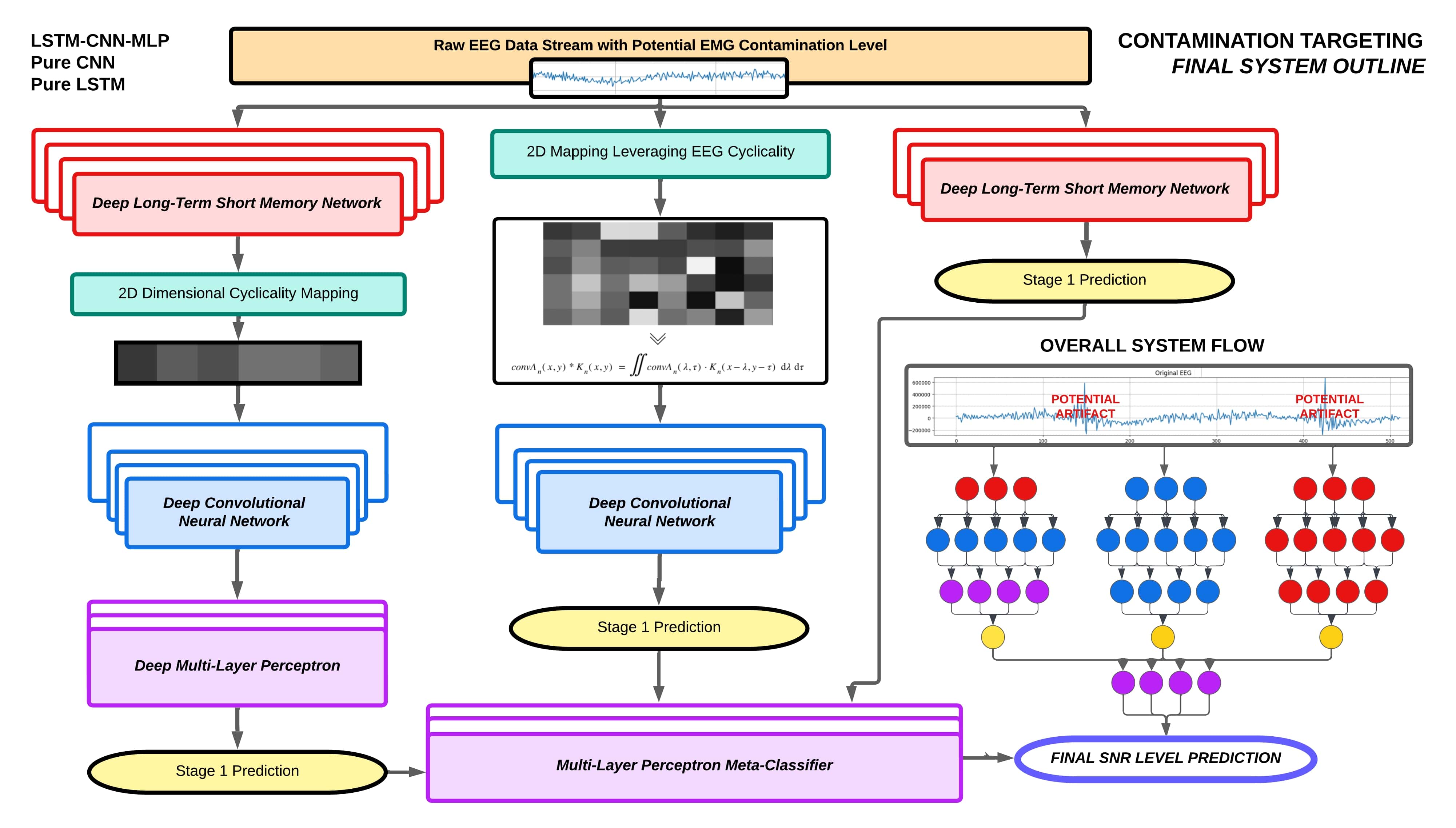}
    \caption{System diagram of the LC ensemble interference detection model. The LSTM-CNN-MLP, CNN, and LSTM ensemble components are depicted from left to right. The MLP meta-classifier model (bottom) is used to aggregate the individual model predictions and output the final SNR level prediction.}
    \label{fig:lc}
\end{figure*}

To provide a comprehensive evaluation of the TADA system's performance, additional quantitative and qualitative analyses were conducted beyond the figures presented in the main text. This section elaborates on these supplementary findings.

Figure~\ref{fig:distribution} illustrates the distribution of TADA test case performance in terms of key performance metrics—Pearson Correlation Coefficient (CC), Temporal Relative Root Mean Squared Error (TRRMSE), and Spectral Relative Root Mean Squared Error (SRRMSE)—across low (-7 dB), mid (-2.5 dB), and high (2 dB) SNR levels. By analyzing the variance of the test case performance, we can observe that the single-segment performance discrepancy between TADA and other top DL models falls within a 90\% error interval (judged by approximate standard deviation) across all metrics—with the sole exception of TRRMSE at -2.5 dB, for which GCTNet’s performance is roughly 2.3 standard deviations above the TADA mean (Yin et al., 2023). TADA represents a $>$90\% reduction in model size compared to the nearest published DL equivalent (and trains significantly faster), so there are intuitive performance tradeoffs.

Zhang et al., 2021b also introduced several denoising benchmark algorithms in conjunction with the dataset publication: two conventional algorithms (empirical mode decomposition and band pass filtration) and four deep learning algorithms (fully-connected neural network, simple convolutional neural network, complex convolutional neural network, and a recurrent neural network). We refer readers to Zhang et al., 2021b for more details, but the recurrent neural network (RNN) generally outperforms the other benchmark algorithms. Compared to the published EEGdenoiseNet benchmarks described in Zhang et al., 2021b, the TADA system largely overperforms all six systems (two conventional, four DL) across the three standard quantitative metrics. The main exception to TADA’s overperformance is at mid SNR (-2.5 dB), where the benchmark RNN model outperformed TADA on the order of roughly a 5\% error reduction across all three metrics. TADA notably outperforms the top-performing RNN benchmark in terms of correlation coefficient by 0.14 at the -7 dB SNR lower bound; thus, while not dominant, TADA remains competitive in terms of SRRMSE, TRRMSE, and CC compared to the major EEGdenoiseNet performance benchmarks. While both adversarial training and the targeting system introduce model complexity, these additions are justified by the aforementioned ablation experiments and performance improvements where applicable.

Further qualitative assessments were undertaken to evaluate TADA system performance edge cases. Figure~\ref{fig:cases} showcases examples of best- and worst-case filtration. As mentioned previously, the most significant improvements in pure MSE and CC noise reduction were observed at lower SNRs (e.g., -7 dB); notable individual reconstructive CCs exceeding 0.85 were achieved at an initial SNR of -7 dB. Substantial spectral and temporal MSE reductions were also broadly evident at low and mid-range initial SNRs, demonstrating the efficiency of logistic covariance-driven rescaling and the robustness of edge-case handling mechanisms integrated with the autoencoder. At high initial SNRs, MSE reduction was less pronounced, which is expected since high starting SNR inherently limits further filtration-driven improvement. However, TADA maintained beneficial signal adjustments at high SNRs without exhibiting degradative behavior, with the aforementioned quantitative results confirming its strong performance even under these conditions.

\subsection{Architectural Details}

The TADA system integrates multiple neural network architectures to achieve efficient and effective EEG time series denoising. This section provides more detailed descriptions of the core architectural components.

The multi-stage LC model used in the meta-targeting SNR inference process integrates LSTM layers, which can capture temporal dependencies in EEG signals, with convolutional layers that are tailored toward spatial feature extraction. The ensemble framework of the LC model is based on prior work (Choi \& Liu, 2025) which demonstrated its effectiveness for EEG time series processing. Together, the ensemble components enable the system to accurately assess the level of EMG interference in incoming EEG data by analyzing both temporal and spatial characteristics of the signal. Driven by the importance of establishing an airtight targeting architecture for our initial TADA demonstration, we fine-tuned our LC model until it was able to fully solve a complete set of target test cases—but preliminary back-of-the-envelope calculations based on LC learning trajectory across the LC model’s 100 epochs raise the possibility for a significant reduction in overall TADA training time with just a 5\% softening of the solvability requirement (LC achieves accuracies north of 95\% after just 7 epochs). Moreover, by expanding the regressive SNR targeting capabilities of the LC model, the system can iteratively smooth remaining SNR performance discontinuities, thereby increasing both overall performance and subsequent tolerance for meta-targeting inaccuracy. The overall architecture of the LC model ensemble is depicted in Figure 7.

The denoising autoencoder (Figure 8) employs a convolutional structure and is comprised of both an encoder and decoder. The encoder processes the noisy time series through two convolutional layers, with 32 and 64 filters, respectively, each followed by batch normalization and ReLU activation. Max-pooling layers with a pool size of 2 are used after each convolutional block to progressively reduce the temporal resolution. The latent representation is generated through a convolutional layer with 128 filters, followed by batch normalization and ReLU activation, capturing essential features of the input. The decoder reconstructs the signal by reversing the encoder's compression process. It employs two convolutional layers with 64 and 32 filters, respectively, each followed by batch normalization, ReLU activation, and upsampling layers to restore the temporal resolution. The final reconstruction is performed by a single convolutional layer with one filter and a sigmoid activation function, ensuring the output has the same shape as the input. The AE is trained using the Adam optimizer and the aforementioned domain-specific loss function.

\begin{figure*}[t]
    \centering
    \includegraphics[width=0.9\textwidth]{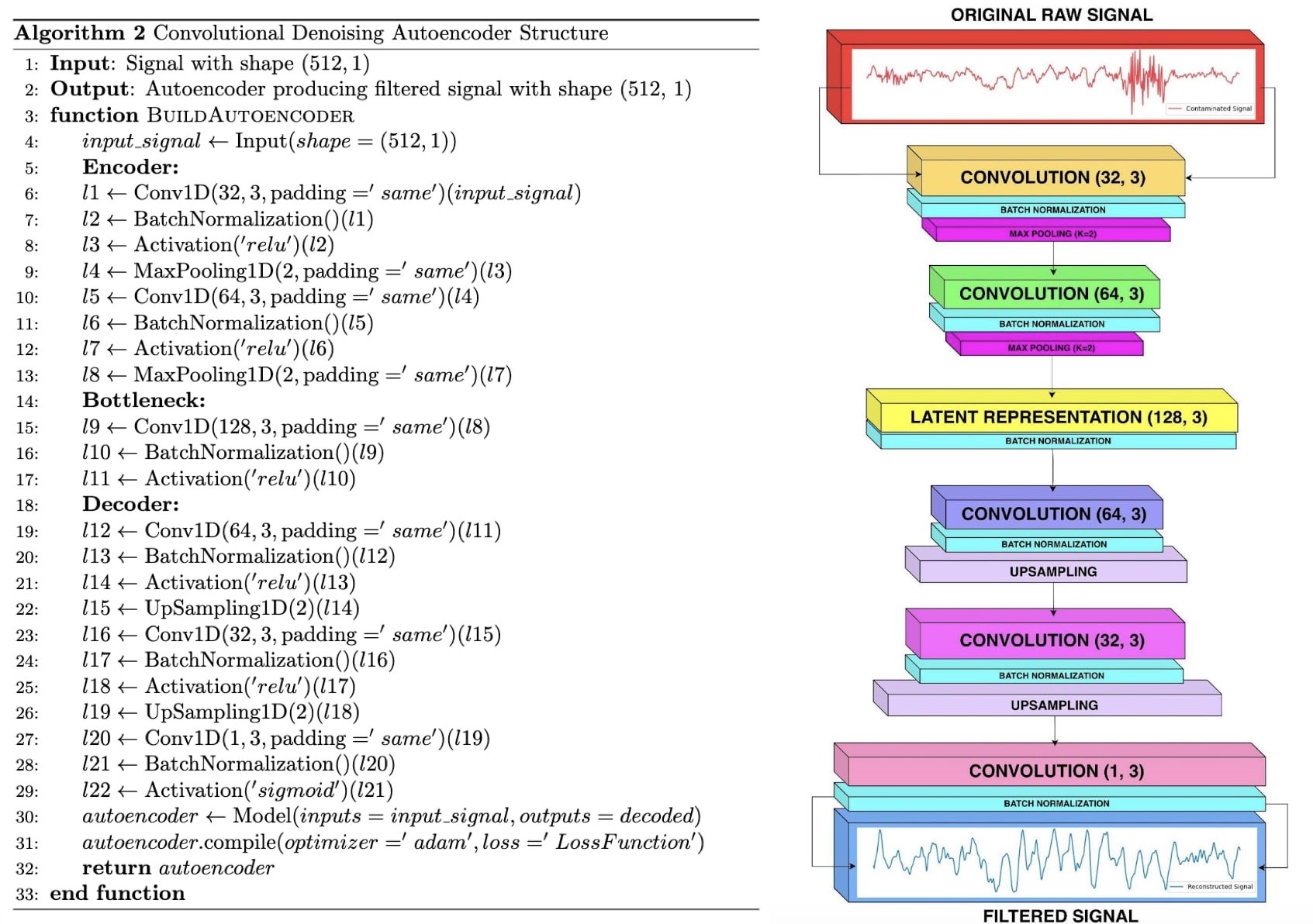}
    \caption{Convolutional denoising autoencoder architecture.}
    \label{fig:gan}
\end{figure*}

\begin{figure*}[!htbp]
    \centering
    \includegraphics[width=0.91\textwidth]{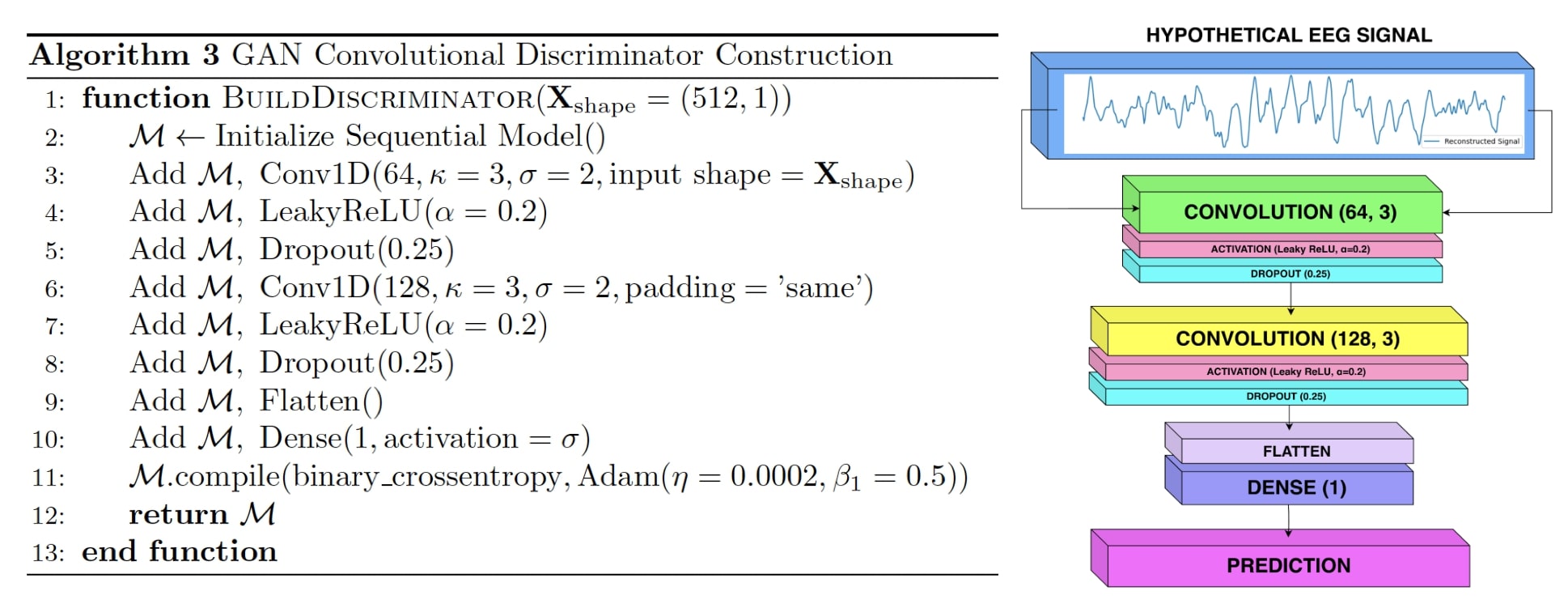}
    \caption{Convolutional discriminator architecture for adversarial autoencoder training.}
    \label{fig:gan}
\end{figure*}

As mentioned previously, the adversarial training environment was implemented by creating an adversarial discriminator model to discern between autoencoder filter-passed samples and real EEG data. The addition of adversarial training in conjunction with the denoising autoencoder custom loss function was hypothesized to elevate the “realism” of the autoencoder output and help minimize signal degradation. The final convolutional discriminator model is designed to differentiate between real (ground truth) and generated (autoencoder output) EEG signals as part of the adversarial training environment via a binary cross-entropy loss function. It consists of sequential convolutional layers with increasing filter sizes, each followed by LeakyReLU activation functions and dropout layers to prevent overfitting; further details are provided in Figure 9. The final layer is a dense layer with a sigmoid activation function, outputting a probability score that indicates the likelihood of the input being a real EEG signal. During training, the discriminator is optimized to accurately classify real and fake signals, thereby providing an adversarial signal that guides the autoencoder towards generating denoised EEG outputs with realistic temporal and spectral properties.

\end{document}